\documentclass{aip-cp}

\usepackage[square, numbers, comma, sort&compress]{natbib}
\usepackage{fancybox}
\usepackage{shortvrb}
\MakeShortVerb\|

\usepackage{listings}
\usepackage[usenames]{color}
\definecolor{codecolor}{gray}{.9}
\definecolor{rlcolor}{cmyk}{0,1,0,0}
\usepackage{amsmath}

\begin{document}

\title{Can a CNN Recognize Catalan Diet?}

\author{Pedro Herruzo}
\eaddress{pherrusa7@alumnes.ub.edu}

\author{Marc Bola\~nos}
\eaddress{marc.bolanos@ub.edu}

\author{Petia Radeva}
\eaddress{petia.ivanova@ub.edu}

\affil{Universitat de Barcelona. Barcelona, Spain.}
\affil{Computer Vision Center. Bellaterra, Spain.}

\maketitle

\begin{abstract}
Nowadays, we can find several diseases related to the unhealthy diet habits of the population, such as diabetes, obesity, anemia, bulimia and anorexia. In many cases, these diseases are related to the food consumption of people. Mediterranean diet is scientifically known as a healthy diet that helps to prevent many metabolic diseases. In particular, our work focuses on the recognition of Mediterranean food and dishes. The development of this methodology would allow to analise the daily habits of users with wearable cameras, within the topic of lifelogging. By using automatic mechanisms we could build an objective tool for the analysis of the patient's behaviour, allowing specialists to discover unhealthy food patterns and understand the user's lifestyle.

With the aim to automatically recognize a complete diet, we introduce a challenging multi-labeled dataset related to Mediterranean diet called FoodCAT. The first type of label provided consists of 115 food classes with an average of 400 images per dish, and the second one consists of 12 food categories with an average of 3800 pictures per class. This dataset will serve as a basis for the development of automatic diet recognition. In this context, deep learning and more specifically, Convolutional Neural Networks (CNNs), currently are state-of-the-art methods for automatic food recognition. In our work, we compare several architectures for image classification, with the purpose of diet recognition. Applying the best model for recognising food categories, we achieve a top-1 accuracy of 72.29\%, and top-5 of 97.07\%. In a complete diet recognition of dishes from Mediterranean diet, enlarged with the Food-101 dataset for international dishes recognition, we achieve a top-1 accuracy of 68.07\%, and top-5 of 89.53\%, for a total of 115+101 food classes.
\end{abstract}

%%%%%%%%%%%%%%%%%%%%%%%%%%%%%%
\section{INTRODUCTION} \label{motivation} 
Technology that helps track health and fitness is on the rise, in particular, automatic food recognition is a hot topic for both, research and industry. People around us have at least 2 devices, such as tablets, computers, or phones, which are used daily to take pictures. These pictures are commonly related to food; people upload dishes to social networks such as Instagram, Facebook, Foodspotting or Twitter. They do it for several reasons, to share a dinner with a friend, to keep track of a healthy diet or to show their own recipes. This amount of pictures is really attractive for companies, who are already putting much effort to understand people's diet, in order to offer personal food assistance and get benefits. 
	
Food and nutrition are directly related to health. Obesity, diabetes, anemia, and other diseases, are all closely related to food consumption. Looking at food habits, the Mediterranean diet is scientifically known as a healthy diet. For example, a growing number of scientific researches has been demonstrating that olive oil, operates a crucial role on the prevention of cardiovascular and tumoral diseases, being related with low mortality and morbidity in populations that tend to follow a Mediterranean diet \cite{2014arXiv1401.2413M}. Many doctors tell patients to write a diary of their diet, trying to make them aware of what they are eating. Usually people do not care too much about that, annotating all the meals often is getting boring. An alternative is to make the food diary by pictures with the phone, or even better, to take the pictures automatically with a small wearable  camera. It can be very useful in order to analyse the daily habits of users with wearable cameras. It appears as an objective tool for the analysis of patient's behaviour, allowing specialists to discover unhealthy food patterns and understand user's lifestyle. However, automatic food recognition and analysis are still challenges to solve for the computer vision community.

\begin{figure}[!ht]
  \centering
  \includegraphics[width=0.6\textwidth]{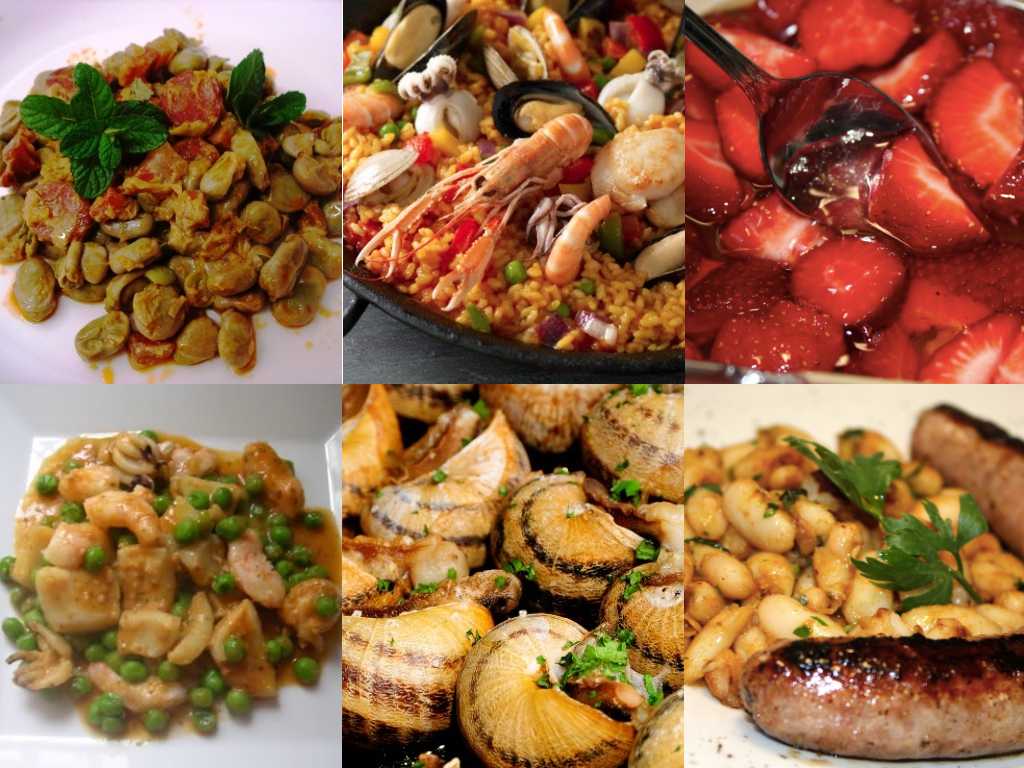} 
  \caption{Examples of Catalan cuisine in \textit{FoodCAT} dataset: sauteed beans, paella, strawberries with vinegar, cuttlefish with peas, roasted snails and beans with sausage.}
  \label{foodCAT}
\end{figure}

Deep learning and more specifically, Convolutional Neural Networks (CNNs) are actually the technologies within the state-of-the-art for automatic food recognition.  The  \textit{GoogleNet} \cite{DBLP:journals/corr/SzegedyLJSRAEVR14} was responsible for setting the state of the art for classification and detection in the ImageNet Large-Scale Visual Recognition Challenge in 2014 \textit{ILSVRC14} \cite{ILSVRC15}. Another widely used model is \textit{VGG} \cite{Simonyan14c}, which secured the first and the second places also for the ImageNet \textit{ILSVRC14} competition \cite{ILSVRC15}, in the localization and classification tasks respectively. One of the most popular food  dataset is the \textit{Food-101} dataset \cite{bossard14}, containing 101 food categories, with 101.000 images. Another well known is  the \textit{UEC FOOD 256} dataset \cite{kawano14c}, which contains 256 types of food. Many researchers have been working with these datasets achieving very good results on food recognition \cite{tatsuma2016food}, or in both food localisation and recognition \cite{DBLP:journals/corr/BolanosR16} \cite{DBLP:conf/icmcs/MatsudaHY12}. Another food related classification task that we are interested in, is to classify food categories, e.g. we should be able to classify a paella picture  into the category of  rice. In our case, we will do it following a robust classification of Catalan diet proposed in the book \textit{El Corpus del patrimoni culinari catal\`a } \cite{corpus}. Other related works on that topic classify 85 food classes \cite{DBLP:conf/ism/HoashiJY10} or 50 dishes \citep{Joutou:2009:FIR:1818719.1818816}. Hence, we construct our dataset from the Catalan cuisine as a good representative of the Mediterranean food.

In this paper we focus on developing automatic algorithms to recognize Catalan food using deep learning techniques. For this purpose we build  a dataset and  enlarge it with the public domain dataset \textit{Food-101}. Our work is organized in three steps:
%We claim to define robust approaches to recognize the datasets with the state of the art methods.

\textit{1. Build a dataset including healthy food}: 
The current food datasets are built in order to achieve a good performance in the general challenge of recognizing pictures automatically. Our goal is to present a method for food recognition of extended dataset based on Catalan food, as it is scientifically supported as a healthy diet (see Fig. \ref{foodCAT} for some examples). Therefore, we present a new dataset based on Catalan food, which we call \textit{FoodCAT}. This dataset has been classified following two different approaches. On one side, the images have been classified based on dishes, and on the other side, in a more general food categories. As an example, our system will recognize a dish with chickpeas with spinach as the food class 'chickpeas with spinach', but also as food category 'Legumes'.

\textit{2. Recognize food dishes with Convolutional Neural Networks}: 
We are interested in applying a Convolutional Neural Network to recognize the new built healthy dataset together with the dataset \textit{Food-101} \cite{bossard14}. We use pre-trained models over the large dataset \textit{ImageNet}, such as \textit{GoogleNet} \cite{DBLP:journals/corr/SzegedyLJSRAEVR14} and the VGG \cite{Simonyan14c}. Moreover, in order to recognize food categories, we compare the differences between fine-tuning a pre-trained model over all the layers, versus the same model trained only for the last fully-connected layer.

\textit{3. Improve the quality of the dataset and the recognition task with Super-Resolution}:
It has been proven that large image resolution improves recognition accuracy \cite{DBLP:journals/corr/WuYSDS15}. Therefore, we will base on a new method to increase the resolution of the images, based on a Convolutional Neural Network, known as Super-Resolution (SR) \cite{wang2015deep}. With that, our goal is to get a better performance in the image recognition task.

% In order to validate our results, we explore different methods such as, balancing all classes for both datasets, or reducing the image resolutions of \textit{Food-101}, making it more similar to the resolutions of \textit{FoodCAT}.
% Finally, we  compare all methods and we will choose the one which gets a better performance.

%%%%% new SECTION
\section{METHODOLOGY}
The image classification problem is the task of assigning a label from a predefined set of categories to an input image. In order to tackle this task for the Catalan diet problem, we propose taking a data-driven approach. After collecting a dataset for the problem at hand, we are going to train a CNN for automatically learning the appearance of each class and classifying them. 

The collected dataset, named \textit{FoodCAT}, when compared to the most widely used dataset for food classification \textit{Food-101}, presents a lower image resolution %(because we collected raw images from Google queries) 
which, as we prove in our experiments, leads to a data bias and a lower performance when training a CNN on the combined datasets. %the \textit{Food-101} dataset: 
%1) it has approximately 400 images per class instead of 1000, and the image resolutions 
 In order to solve this problem, 
 %which can be usually found on images collected from the internet, 
 we must increase the resolution to at least 256x256 pixels, which is the usual input size to CNNs. Thus, we propose using the method known as \textit{Super-Resolution} and consequently improve the accuracy in the food recognition task.

% new subsection
\subsection{Model} \label{models}

In order to apply food classification, we propose using the \textit{GoogleNet} architecture, which has proven to obtain very high performance in several classification tasks \cite{tatsuma2016food} \cite{DBLP:journals/corr/BolanosR16} \cite{2016arXiv160705440J}.
%and, as we have proven (see section \ref{sec:exp_results}), it works better than other approaches for our problem.

We train the \textit{GoogleNet} model using an image crop of 224x224x3 pixels as input. During training, in order to perform data augmentation, we extract random crops from the images after unifying their resolution to 256x256x3. During the testing procedure, we use the central image crop. The \textit{GoogleNet}  convolutional neural network architecture is a replication of the model described in the \textit{GoogleNet} publication \cite{DBLP:journals/corr/SzegedyLJSRAEVR14}. The network is 22 layers deep when counting only layers with parameters (or 27 layers if we also count pooling layers). As the authors explain in their paper \cite{DBLP:journals/corr/SzegedyLJSRAEVR14}, two of the features that made this net so powerful are :
\textit{Auxiliary classifiers connected to the intermediate layers}: which was thought to combat the vanishing gradient problem given the relatively large depth of the network. During training, their loss gets added to the total loss of the network with a discount weight. In practice, the auxiliary networks effect is relatively minor (around 0.5\%) and it is required only one of them to achieve the same effect.
\textit{Inception modules}: the main idea for it is that in images, correlations tend to be local. Therefore, in each of the 9 modules, they use convolutions of dimension 1x1, 3x3, 5x5, and pooling layers of 3x3. Then, they put all outputs together as a concatenation. Note that to reduce the depth of the volume, convolutions 3x3 and 5x5 are performed after applying a 1x1 convolution with less filters, and pooling 3x3 is also followed by a convolution 1x1. This  makes the model more efficient reducing the number of parameters in the net.

%\textcolor{red}{Marc: if I am not wrong GoogleNet was the best model for our problem. Considering that in the current section we should only explain our complete best proposal, we should move the following explanation of the VGG to the section where we explain the experimental setup. pEDRO: If we will move it there, should we start another subsection? As we already mentioned the}

%\textbf{VGG-19}: This net has 5 blocks of different depth convolutions (64, 128, 256, 512, and 512 consecutively) and 3 FC layers. The first 2 blocks contains 2 different convolutions each and the last 5 contains 4 different convolutions each. It is a total of $2\times2+3\times4+3=19$ layers. All convolutions have a kernel size of 3x3 with a padding of 1 pixel, i.e. the spatial resolution is preserved after convolution. Finally, after each convolutional block a max pooling is performed over a 2x2 pixel window, with stride 2, i.e. reducing by a factor of 2 the spatial size after each block. Section 2.3 in VGG-19 paper \cite{Simonyan14c}, confirm that small-size convolution filters are the key, together with apply deep CNN, to outperform googleNet in ILSVRC14 \cite{ILSVRC15} in terms of the single-network classification accuracy. 

% new subsection
\subsection{Super-Resolution} \label{SR}

The image dimensions of \textit{FoodCAT} dataset are on average smaller than 256x256. 
%Therefore, the images are increased to this size as an input for Caffe, the Deep Learning framework that we are using, causing deformation and noise. We found interesting . 
Motivated by the fact that larger images improve recognition accuracy \cite{DBLP:journals/corr/WuYSDS15}, we propose increasing the resolution with a state-of-the-art method instead of applying a common upsampling through bilinear interpolation. % leave the net do it with the regular resize, the image recognition accuracy is better.
 To increase the size of the images, we use the method called Super-Resolution \cite{wang2015deep}. In this paper, the authors propose a technique for obtaining a High-Resolution (HR) image from a Low-Resolution (LR) one. To this end, they use a Sparse Coding based Network (SCN) based on the Learned Iterative Shrinkage and Thresholding Algorithm (LISTA) \cite{DBLP:conf/icml/2010}. Notable improvements are achieved over the generic CNN model in terms of both recovery accuracy and human perception. The implementation is based on recurrent layers that merge linear adjacent ones, allowing to jointly optimize all the layer parameters from end to end. It is achieved by  rewriting the activation function of the LISTA layers as follows:
\begin{equation*}
[h_\theta (a)]_i = sign(a_i )\theta_i (\|{a_i}\|/\theta_i − 1)_+ = \theta_i h_1 (a_i /\theta_i )
\end{equation*}

\begin{figure}[!ht]
  \centering
  \includegraphics[width=0.3\textwidth]{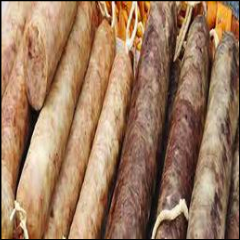} \;\;
  \includegraphics[width=0.3\textwidth]{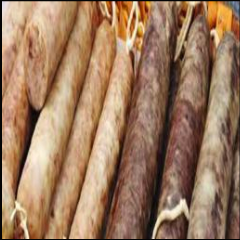}
  \caption{Left shows the SR decreased to 256x256 and right shows the original increased to 256x256.} 
  \label{SRfoodCAT}
\end{figure}

% \textcolor{red}{\textbf{Marc}: I think that in this section we should explain how does the SR method work. As Petia said we should include some formula or some more detailed explanation of how the method works. Additionally, the following figures and tables as well as the following part of text would be more suited to the results section in order to justify the output of the SR method.
% \textbf{Pedro}: I did. Yes, we can move it}

Fig. \ref{SRfoodCAT} shows the visual difference of a randomly chosen FoodCAT image compared to its SR version. In this example, the original image is 402x125, so the SR was applied with a factor of 3 to assure that both dimensions are bigger than 256.

%%%%% new SECTION
\section{RESULTS}
In this section, we describe the datasets, metrics used for evaluating and comparing each model, and results for each of the image recognition tasks: dishes and food categories.

\subsection{Dataset} \label{categories}
Our dataset, \textit{FoodCAT} has two different labels for each image: Catalan dish, and Catalan food category. Although the total number of Catalan dishes of our datasets are 140, we selected only the set of classes with at least 100 images for our experiments, resulting in a total of 115 classes. Some examples of the available dishes are: sauteed beans, paella, strawberries with vinegar, cuttlefish with peas, roasted snails or beans with sausage. In addition, the images are also labeled in 12 general food categories. Table \ref{table:classes} shows a summary of the general statistics of the dataset, including the number of dishes and images that we have tagged for each food category.  

\begin{table}[!htbp]
\centering
\begin{tabular}{lllll}
\cline{1-4}
 & \textbf{\# dishes} & \textbf{\%} & \textbf{\# images} \\ \hline
\multicolumn{1}{l}{Desserts and sweets} & 34 & 24,28 & 11.933 \\ \hline
\multicolumn{1}{l}{Meats} & 26 & 18,57 & 7.373 \\ \hline
\multicolumn{1}{l}{Seafood} & 25 & 17,85 & 5.977 \\ \hline
\multicolumn{1}{l}{Pasta, rice and other cereals} & 11 & 7,85 & 4.728 \\ \hline
\multicolumn{1}{l}{Vegetables} & 11 & 7,85 & 3.007 \\ \hline
\multicolumn{1}{l}{Salads and cold dishes} & 5 & 3,57 & 2.933 \\ \hline
\multicolumn{1}{l}{Soups, broths and creams} & 8 & 5,71 & 2.857 \\ \hline
\multicolumn{1}{l}{Sauces} & 4 & 2,85 & 2.462 \\ \hline
\multicolumn{1}{l}{Legumes} & 6 & 4,28 & 1.920 \\ \hline
\multicolumn{1}{l}{Eggs} & 5 & 3,57 & 615 \\ \hline
\multicolumn{1}{l}{Snails} & 3 & 2,14 & 470 \\ \hline
\multicolumn{1}{l}{Mushrooms} & 2 & 1,42 & 438 \\ \hline \hline
\multicolumn{1}{l}{\textbf{Total}} & 140 & 100 & 44.713 \\ \hline
\end{tabular}
\caption{First column lists the categories, second and third column show the number and the percentage of dishes, and the fourth one shows the amount of pictures by category.}
\label{table:classes}
\end{table}

\subsection{Implementation}
There are several frameworks with high capabilities for working on the field of Deep Learning such as TensorFlow, Torch, Theano, Caffe, Neon, etc. We choose Caffe, because it tracks the state-of-the-art in both code and models and is fast for developing. We also decided to use it, because it has a large community giving support on the Caffe-users group and Github, uploading new pre-trained models that people can use for different purposes. %It allows a faster training and better accuracy \cite{2014arXiv1405.3531C}, because normally these models are trained originally for larger datasets that the ones, we will use on the fine-tuning. 

A competitive alternative of the \textit{GoogleNet} model is the \textit{VGG-19}, which we also use in our experiments. This net has 5 blocks of different depth convolutions (64, 128, 256, 512, and 512 consecutively) and 3 FC layers. The first 2 blocks contain 2 different convolutions each and the last 5 contain 4 different convolutions each. It has a total of $2\times2+3\times4+3=19$ layers. All convolutions have a kernel size of 3x3 with a padding of 1 pixel, i.e. the spatial resolution is preserved after each convolution. Finally, after each convolutional block a max pooling is performed over a 2x2 pixel window with stride 2, i.e. reducing by a factor of 2 the spatial size after each block. As the VGG-19 paper \cite{Simonyan14c} shows,  small-size convolution filters are the key to outperform the GoogleNet in ILSVRC14 \cite{ILSVRC15} in terms of the single-network classification accuracy.

\subsection{Evaluation Metrics}
Many metrics can be considered to measure the performance of a classification task. In the literature, mainly three methods are used: Accuracy {Top-1} (AT1), Accuracy {Top-5} (AT5), and the Confusion Matrix (CM). In real-world applications, usually the dataset contains unbalanced classes and the above measures can hide the misclassification of classes with fewer samples. Hence, we consider the Normalized Accuracy {Top-1} (NAT1), that gives us the information of how good the classifier is no matter how many samples each class has. 
Let us define formally each metric.

Let $N$ be the total number of classes with images to test, let $N_{i}$ be the number of images of the $i$-th class, and set $n=\sum_{i=0}^{N-1} N_{i}$, as the total number of images to test. Let $\hat{y}_{i,j}^k$ be the $\texttt{top-k}$ predicted classes of the $j$-th image of the $i$-th class, and $y_{i,j}$ the corresponding true class.
Let us also define $\mathbf {1} _{A}\colon X\to \{0,1\}\,$ as the indicator function as follows:
\begin{equation*}
{\displaystyle \mathbf {1} _{A}(x):={\begin{cases}1&{\text{if }}x_{i}\in A \text{, for some } i,\\0&{\text{if }}x_{i}\notin A \text{, for all } i.\end{cases}}}
\end{equation*}

Then, the definitions of the metrics are as follows:

\begin{equation*}
\texttt{AT1} = \frac{1}{n} \sum_{i,j}{1} _{y_{i,j}}(\hat{y}_{i,j}^1), \;\;\;\;
\texttt{AT5} = \frac{1}{n} \sum_{i,j}{1} _{y_{i,j}}(\hat{y}_{i,j}^5), \;\;\;\;
\texttt{NAT1} = \frac{1}{N} \sum_{i=0}^{N-1} \frac{1}{N_{i}} \sum_{j=0}^{N_{i}-1} {1} _{y_{i,j}}(\hat{y}_{i,j}^1).
\end{equation*}

\subsection{Super Resolution application}
\begin{figure}[!ht]
  \centering
  \includegraphics[width=0.3\textwidth]{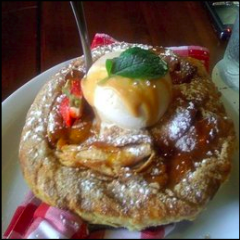} \;\;
  \includegraphics[width=0.3\textwidth]{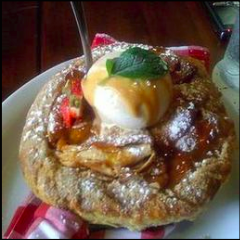}
  \caption{Example of SR used in a high resolution image. Left: original image 512x512 resized to 256x256. Right: original image reduced at 40\% 230x230, then increased by the SR two times to 460x460, and finally resized to 256x256.} 
  \label{SRfood-101}
\end{figure}

\begin{figure}[!ht] 
  \centering
  \includegraphics[width=0.75\textwidth]{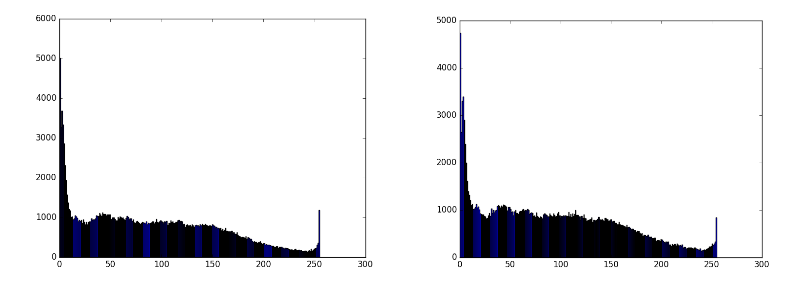}
  \caption{Histograms of   the original image (left), and the SR (right).} 
  \label{fig:hist}
\end{figure}

For all \textit{FoodCAT} images, we applied the SR method in order to make both image dimensions, width and height, bigger or equal to 256.
%\subsubsection{Examples of SR for images in the used datasets}
%\textbf{Example with Food-101}:
In Fig. \ref{SRfood-101}, we show the behaviour of the SR algorithm applied on a Food-101 image. On the left, we show the original image (512x512) resized to the network's input 256x256, and on the right, we show the same image after resizing it to a smaller resolution than the network's input and applying the SR method for also obtaining a results of 256x256. Thus, we simulate the result of the SR procedure on FoodCAT images: first, improvement through SR and second, resizing to the network's input. 
We can see that, from a human perception perspective, applying the SR to a low resolution image does not affect the result. Also, when computing %standard image measures, as the mean, min and max pixel value (see Table \ref{tb:SRfood-101}) and plotting 
the histogram of both images (see Fig. \ref{fig:hist}), one can see that the difference between them is negligible.

%\begin{table}[!ht]
%\caption{Mean, max, and min computed for two images. Original: original image resized to 256x256. SR: reduced at 40\% 230x230, then increased by SR x2 to 460x460, and finally resized to 256x256.}
%\label{tb:SRfood-101}
%\centering
%\begin{tabular}{ccccccc}
%\hline
% & & \textbf{mean} & & \textbf{max} & &\textbf{min} \\ \hline
%\multicolumn{1}{l}{Original} & & 91.10 & & 255 & & 91.10 \\ \hline
%\multicolumn{1}{l}{SR} & & 91.04 & & 255 & & 91.04 \\ \hline
%\end{tabular}
%\end{table}

\subsection{Experimental Results} \label{sec:exp_results}
\begin{figure}[!ht]
\resizebox{\textwidth}{!}{
  %\hspace*{-2.5cm}
  \centering
  \includegraphics[width=1\textwidth]{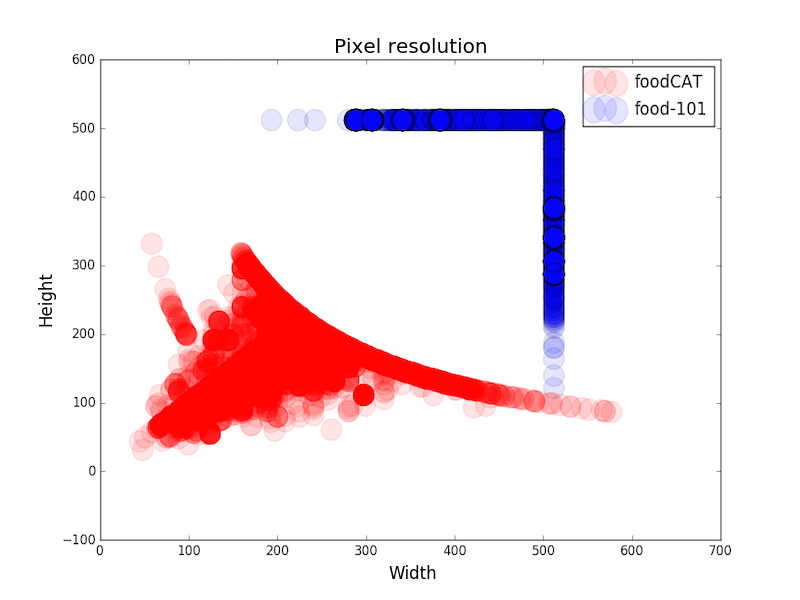}
  \includegraphics[width=1\textwidth]{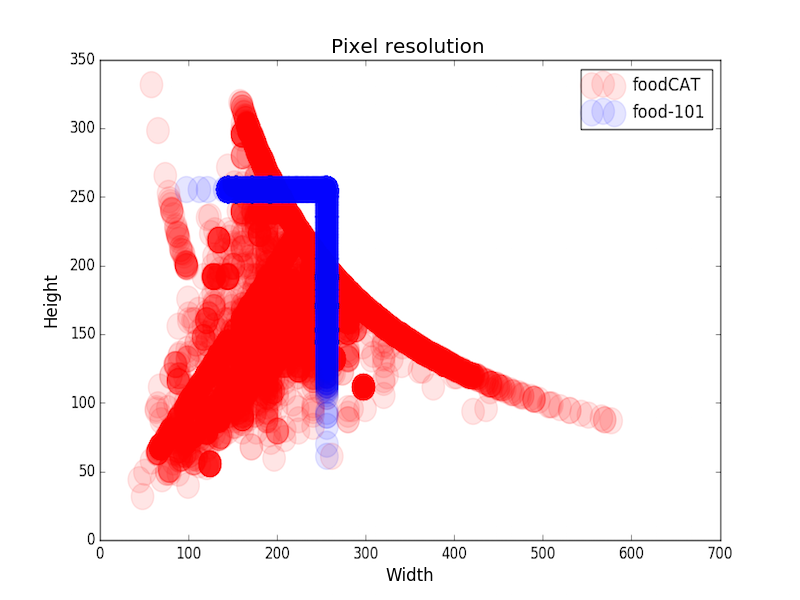}
  \includegraphics[width=1\textwidth]{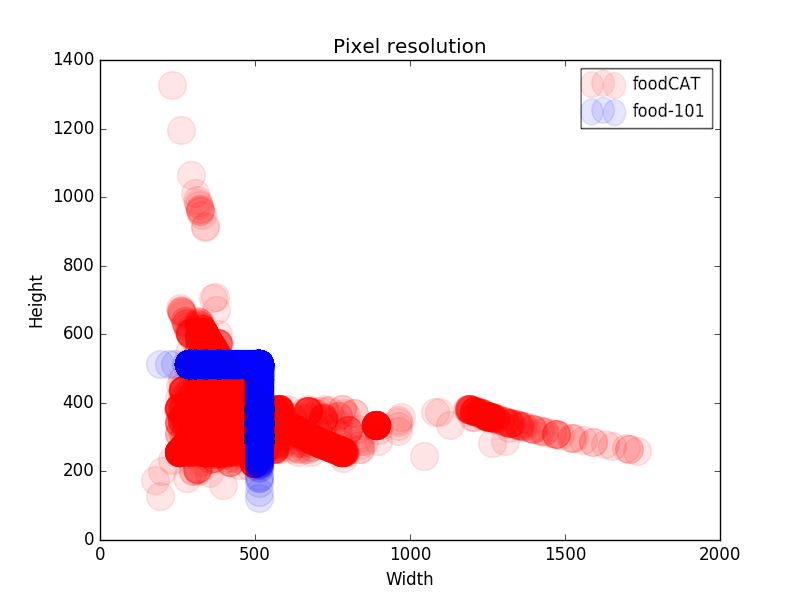} 
  }
  \caption{Plots of image dimension distributions: left: \textit{Food-101}+\textit{FoodCAT}; center: \textit{Food-101} halved+\textit{FoodCAT} with resolution halved, and right: \textit{Food-101}+\textit{FoodCAT} with SR.}
  \label{combined}
\end{figure}

We need to test the performance of the convolutional neural network on both: dish and food category recognition.
\textbf{Dish recognition}: One of the richest public domain datasets is the \textit{Food-101} dataset. Since there is small intersection of both datasets, we decided to  combine the \textit{FoodCAT}  and the \textit{Food-101} dataset in order to build a joint classification model for several types of food. However, in this case we must deal with the differences in image resolution.
In order to tackle this problem, we compared the classification on three different dataset configurations (see Fig. \ref{combined}).

\textbf{a) \textit{Food-101}+\textit{FoodCAT}}: in this experiment, we use the original images. While all pictures in \textit{Food-101} dataset have similar dimension (width or height) equal to 512, the pictures in \textit{FoodCAT} have a huge diversity in resolutions and do not follow any pattern. On average, their resolution is below 256x256. 

%\textbf{\textit{Food-101}+\textit{FoodCAT}}:
%Left in figure \ref{combined} shows the dimension of all original images for each dataset. We can observe how \textit{Food-101} dataset follows the pattern to have one dimension (width or height) equal to 512, and \textit{FoodCAT} has a huge diversity of resolutions, but in average, lower than 256x256. As the CNN models that we are using requires images with dimension 256x256, images of \textit{FoodCAT} are increased and deformed to 256x256, whereas that images of \textit{Food-101} are decreased and deformed for the training. Thus, we are using images with much more noise in \textit{FoodCAT} than in \textit{Food-101}. The next two different dataset configurations are created in order to face this problem.  

\textbf{b) \textit{Food-101} halved+\textit{FoodCAT}}: in this experiment, we decreased the resolution of all images in \textit{Food-101} to make them more alike \textit{FoodCAT}.%, and add noise to the images.

\textbf{c) \textit{Food-101}+\textit{FoodCAT} with SR}: in this experiment, we increased the resolution of all images in \textit{FoodCAT} with the {SR} technique. Therefore, augmenting the resolution allows to reach a higher fidelity than increasing it with a standard resizing method.

% combined

Another of the problems, we have to deal with, when joining two different datasets is the unbalance of classes. Table \ref{table:datasets} shows the number of images per learning phase either when using all images (top row) or a maximum of 500 images per class for balance (bottom row).

\begin{table}[!h]
\centering
%\hspace*{-3cm} 
\resizebox{\textwidth}{!}{
\begin{tabular}{ccccc}
\cline{1-5}
\multicolumn{1}{l}{} & \textbf{training} & \textbf{validation} & \textbf{testing} & \textbf{total} \\ \hline
\multicolumn{1}{c}{Complete} & 116.248 (80.800+35.448) & 14.540 (10.100+4.440) & 14.516 (10.100+4.416) & 145.304 (101.000+44.304) \\ \hline
\multicolumn{1}{c}{Balanced} & 73.085 (40.400+32.685) & 9.143 (5.050+4.093) & 9.124 (5.050+4.074) & 91.352 (50.500+40.852) \\ \hline
\end{tabular}}
\caption{Number of images per learning phase (training, validation and testing) over the complete dataset and the balanced one. The values are presented giving the total number of images in addition to the relative contribution of each dataset in brackets (\textit{Food-101}+\textit{FoodCAT}).}
\label{table:datasets}
\end{table}

As a result, dish recognition is performed over \textit{FoodCAT} and \textit{Food-101}, having 115+101 classes to classify respectively. We study the network performance depending on image resolutions and balanced/unbalanced classes. The 6 different experiments are listed below, denoting \textit{GoogleNet} as 'G' and \textit{VGG-19} 'V':
\begin{enumerate}
\item G: \textit{Food-101} + \textit{FoodCAT} with SR.
\item G: \textit{Food-101} + \textit{FoodCAT} with SR, all balanced.
\item G: \textit{Food-101} halved + \textit{FoodCAT}.
\item G: \textit{Food-101} halved + \textit{FoodCAT}, all balanced.
\item V: \textit{Food-101} + \textit{FoodCAT}.
\item V: \textit{Food-101} + \textit{FoodCAT}, all balanced.
\end{enumerate}

For all the experiments, we fine-tune our networks after pre-training them on the ImageNet dataset.

Table \ref{exp:1to6} organises the results of all the 6 different experiments applied either on both datasets ('A, B') or on \textit{FoodCAT} only ('B'). We set the best $AT1$, $AT5$, and $NAT1$ in bold, for each of the tested datasets (\textit{Food-101}+\textit{FoodCAT} or \textit{FoodCAT}).
We can see that the best results for the dataset \textit{FoodCAT} (columns 'B') are achieved by a {CNN} trained from the original dataset (without SR) with balanced classes (experiment 6). It shows the importance of the balanced classes to recognize, with similar accuracy, different datasets with a single {CNN}.
Furthermore, the results of the test in both datasets together (columns 'A, B') are better, when we use all samples in both datasets during the training phase with the method SR applied for the \textit{FoodCAT}. This {CNN} is the one used in experiment 1, and it also achieves the second best result for the $AT1$ over the \textit{FoodCAT} dataset, with a score of $50.02$, just $0.57$ less than the balanced datasets with {VGG} (experiment 6). Moreover, adding all scores for the accuracy $AT1$ and $AT5$, over the two tests 'A, B' and 'B', experiment $1$ has the highest value of $289.44$ followed by experiment 6 with value $288.09$.

With all this data, we conclude that the best model is the \textit{GoogleNet} trained from all samples of both datasets, with the SR method applied for \textit{FoodCAT}, corresponding to experiment 1. 
% In our future work, we will train a mix over the two winners for these tests: the VGG model with the SR method applied for \textit{FoodCAT} and both balanced datasets.
% We want to note that the best models (corresponding to experiments 1 and 6), during the training phase,  continued learning until the last iteration, i.e. the last iteration has the best performance. Therefore, if we let the models to learn longer, the nets should get a better performance.

\begin{table} [!htbp]
\resizebox{\textwidth}{!}{\begin{tabular}{lccccccccccll}
\cline{1-13}
\multicolumn{1}{l}{Experiment} & \multicolumn{2}{c}{\textbf{1}} & \multicolumn{2}{c}{\textbf{2}} & \multicolumn{2}{c}{\textbf{3}} & \multicolumn{2}{c}{\textbf{4}} & \multicolumn{2}{c}{\textbf{5}} & \multicolumn{2}{c}{\textbf{6}} \\ \hline
\cline{1-13}
 \multicolumn{1}{l}{Datasets} & \multicolumn{1}{l}{A, B} & \multicolumn{1}{l}{B} & \multicolumn{1}{l}{A, B} & \multicolumn{1}{l}{B} & \multicolumn{1}{l}{A, B} & \multicolumn{1}{l}{B} & \multicolumn{1}{l}{A, B} & \multicolumn{1}{l}{B} & \multicolumn{1}{l}{A, B} & \multicolumn{1}{l}{B} & A, B & B \\ \hline
\multicolumn{1}{l}{$AT1$} & \textbf{68.07} & 50.02 & 62.41 & 48.94 & 67.16 & 49.66 & 61.28 & 48.85 & 67.74 & 48.12 & 65.16 & \textbf{50.59} \\ \hline
\multicolumn{1}{l}{$AT5$} & \textbf{89.53} & 81.82 & 86.81 & 81.63 & 89.27 & 82.07 & 86.52 & 80.92 & 89.28 & 81.03 & 88.94 & \textbf{83.40} \\ \hline
\multicolumn{1}{l}{$NAT1$} & 59.08 & 44.25 & 57.91 & 44.44 & 58.57 & 44.31 & 56.99 & 44.44 & 58.18 & 42.34 & \textbf{60.74} & \textbf{46.53} \\ \hline
\end{tabular}}
\caption{Results of the experiments from 1 to 6. A$=$\textit{Food-101}, B$=$\textit{FoodCAT}.}
\label{exp:1to6}
\end{table}

\textbf{Food categories recognition}:
The recognition of food categories is performed over the \textit{FoodCAT} dataset by fine-tuning the \textit{GoogleNet} CNN trained previously with the large dataset ImageNet. We study the network performance depending on if we train all layers or only the last one, the fully-connected layer. Table \ref{table:categories} shows the results obtained for this task. First, if we have a limited machine or limited time, we show that fine-tuning just the fully-connected layer over a model previously trained on a large dataset as \textit{ImageNet} \cite{imagenet_cvpr09}, it can give a good enough performance.  Training all layers, we achieve recognition of food categories over Catalan food with $AT1=72.29$ and $AT5=97.07$. Taking care of the difference of samples on each class, the normalized measure also gives a high performance, with $NAT1=65.06$.

\begin{table}[!htbp]
\centering
\begin{tabular}{lllllll}
\cline{1-7}
 & \textbf{AT1} & \textbf{AT5} & \textbf{NAT1} & \textbf{\# Iterations} & \textbf{Best iteration} & \textbf{Time executing} \\ \hline
\multicolumn{1}{l}{FC} & 61.36 & 93.39 & 50.78 & 1.000.000 & 64.728 & 12h \\ \hline
\multicolumn{1}{l}{All layers} & \textbf{72.29} & \textbf{97.07} & \textbf{65.06} & 900.000 & 49.104 & 24h \\ \hline
\end{tabular}
\caption{Performance and learning time, fine-tuning the \textit{GoogleNet} model over the food categories labels. We show the results for two experiments done: training all layers, and only training the last fully-connected.}
\label{table:categories}
\end{table}

\begin{figure}[!ht]
  \centering
  \includegraphics[width=0.7\textwidth]{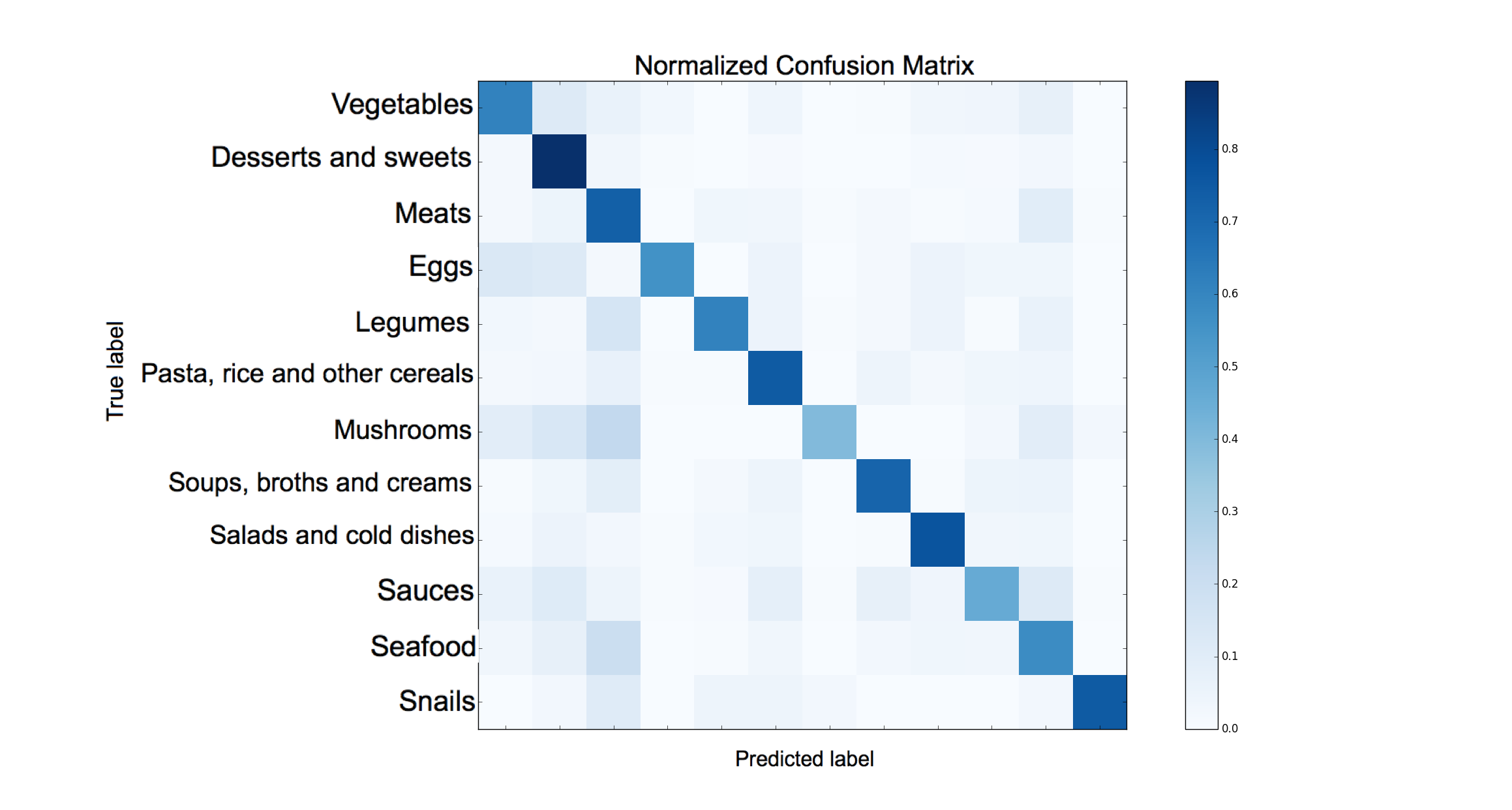}
  \caption{Normalized CM of \textit{GoogleNet} model trained over the all layers to recognize food categories.}
  \label{cmCategories}
\end{figure}

Figure \ref{cmCategories} shows the normalized Confusion Matrix for the \textit{GoogleNet} model trained over all layers. It is not surprising that  'Desserts and sweets' is the category that the net can recognize better, as it is also the class with more samples in the dataset with 11.933 images, followed by 'Meats' with 7.373. We also must note that the classes with less samples in our dataset are 'Snails' and 'Mushrooms', but those specific classes can also be found in the \textit{ImageNet} (the dataset used for the pre-trained model that we are using) that explains the good performance of the network on them.

%%%%% new SECTION
\section{CONCLUSIONS}
In this paper, we presented the novel and challenging multi-labeled dataset related to the Catalan diet called \textit{FoodCAT}. For the first kind of labels, the dataset is divided into 115 food classes with an average of 400 images per dish. For the second kind of labels, the dataset is divided into 12 food categories with an average of 3800 images per dish.

We explored the  food classes recognition and found that the best model  is obtained by fine-tuning the GoogleNet network on the datasets \textit{FoodCAT}, after increasing the resolution with the  Super-Resolution method and \textit{Food-101}. This model achieves the highest accuracy top-1 with 68.07\%, and top-5 with 89.53\%, testing both datasets together, and top-1 with 50.02\%, and top-5 with 81.82\%, testing only \textit{FoodCAT}.
Regarding the food categories recognition, we achieved the highest accuracy top-1 with 72.29\% and top-5 with 97.07\%, after fine-tuning the \textit{GoogleNet} model for all layers. Our next steps are to increase the dataset and explore other architectures of convolutional neural networks for food recognition.

%%%%% new SECTION
\section{ACKNOWLEDGMENTS}
This work was partially funded by TIN2015-66951-C2-1-R, La Marat\'o de TV3, project 598/U/2014 and SGR 1219.
P. Radeva is supported by an \textit{ICREA Academia} grant. Thanks to the University of Groningen for letting us use the Peregrine HPC cluster.

%\noindent\Ovalbox{\verb"\bibliographystyle{style-name}"}\\

%For a AIP Conference Proceddings we recommend the use of
%|aipnum-cp.bst| provided with this packet. This bibliograpgy style
%generates numbered style references in the required format.\\

%\noindent\Ovalbox{\verb"\bibliography{bib-list}"}\\

%% ----------------------------------------------------------------
\label{Bibliography}
\bibliographystyle{unsrtnat}  % Use the "unsrtnat" BibTeX style for formatting the Bibliography
\bibliography{guide}  % The references (bibliography) information are stored in the file named "Bibliography.bib" 

\begin{thebibliography}{17}
\providecommand{\natexlab}[1]{#1}
\providecommand{\url}[1]{\texttt{#1}}
\expandafter\ifx\csname urlstyle\endcsname\relax
  \providecommand{\doi}[1]{doi: #1}\else
  \providecommand{\doi}{doi: \begingroup \urlstyle{rm}\Url}\fi

\bibitem[{Monteiro-Silva}(2014)]{2014arXiv1401.2413M}
F.~{Monteiro-Silva}.
\newblock {Olive oil's polyphenolic metabolites - from their influence on human
  health to their chemical synthesis}.
\newblock \emph{ArXiv e-prints 1401.2413}, January 2014.

\bibitem[Szegedy et~al.(2014)Szegedy, Liu, Jia, Sermanet, Reed, Anguelov,
  Erhan, Vanhoucke, and Rabinovich]{DBLP:journals/corr/SzegedyLJSRAEVR14}
Christian Szegedy, Wei Liu, Yangqing Jia, Pierre Sermanet, Scott~E. Reed,
  Dragomir Anguelov, Dumitru Erhan, Vincent Vanhoucke, and Andrew Rabinovich.
\newblock Going deeper with convolutions.
\newblock \emph{CoRR}, abs/1409.4842, 2014.
\newblock URL \url{http://arxiv.org/abs/1409.4842}.

\bibitem[Russakovsky et~al.(2015)Russakovsky, Deng, Su, Krause, Satheesh, Ma,
  Huang, Karpathy, Khosla, Bernstein, Berg, and Fei-Fei]{ILSVRC15}
Olga Russakovsky, Jia Deng, Hao Su, Jonathan Krause, Sanjeev Satheesh, Sean Ma,
  Zhiheng Huang, Andrej Karpathy, Aditya Khosla, Michael Bernstein,
  Alexander~C. Berg, and Li~Fei-Fei.
\newblock {ImageNet Large Scale Visual Recognition Challenge}.
\newblock \emph{International Journal of Computer Vision (IJCV)}, 115\penalty0
  (3):\penalty0 211--252, 2015.
\newblock \doi{10.1007/s11263-015-0816-y}.

\bibitem[Simonyan and Zisserman(2014)]{Simonyan14c}
K.~Simonyan and A.~Zisserman.
\newblock Very deep convolutional networks for large-scale image recognition.
\newblock \emph{CoRR}, arXiv:1409.1556, 2014.

\bibitem[Bossard et~al.(2014)Bossard, Guillaumin, and Van~Gool]{bossard14}
Lukas Bossard, Matthieu Guillaumin, and Luc Van~Gool.
\newblock Food-101 -- mining discriminative components with random forests.
\newblock In \emph{European Conference on Computer Vision}, 2014.

\bibitem[Kawano and Yanai(2014)]{kawano14c}
Y.~Kawano and K.~Yanai.
\newblock Automatic expansion of a food image dataset leveraging existing
  categories with domain adaptation.
\newblock In \emph{Proc. of ECCV Workshop on Transferring and Adapting Source
  Knowledge in Computer Vision (TASK-CV)}, 2014.

\bibitem[Tatsyma and Masaki(2016)]{tatsuma2016food}
Atsushi Tatsyma and Aono Masaki.
\newblock Food image recognition using covariance of convolutional layer
  feature maps.
\newblock \emph{IEICE TRANSACTIONS on Information and Systems}, 99\penalty0
  (6):\penalty0 1711--1715, 2016.

\bibitem[Bola{\~{n}}os and Radeva(2016)]{DBLP:journals/corr/BolanosR16}
Marc Bola{\~{n}}os and Petia Radeva.
\newblock Simultaneous food localization and recognition.
\newblock \emph{In Proceedings of the International Conference on Pattern
  Recognition (in press)}, 2016.
\newblock URL \url{http://arxiv.org/abs/1604.07953}.

\bibitem[Matsuda et~al.(2012)Matsuda, Hoashi, and
  Yanai]{DBLP:conf/icmcs/MatsudaHY12}
Yuji Matsuda, Hajime Hoashi, and Keiji Yanai.
\newblock Recognition of multiple-food images by detecting candidate regions.
\newblock In \emph{Proceedings of the 2012 {IEEE} International Conference on
  Multimedia and Expo, {ICME} 2012, Melbourne, Australia, July 9-13, 2012},
  pages 25--30, 2012.
\newblock \doi{10.1109/ICME.2012.157}.
\newblock URL \url{http://dx.doi.org/10.1109/ICME.2012.157}.

\bibitem[de~la Cuina(2011)]{corpus}
Institut~Catal\'{a} de~la Cuina.
\newblock \emph{Corpus del patrimoni culinari catal\'{a}}.
\newblock Edicions de la Magrana, 2011.
\newblock ISBN 9788482649498.

\bibitem[Hoashi et~al.(2010)Hoashi, Joutou, and
  Yanai]{DBLP:conf/ism/HoashiJY10}
Hajime Hoashi, Taichi Joutou, and Keiji Yanai.
\newblock Image recognition of 85 food categories by feature fusion.
\newblock In \emph{12th {IEEE} International Symposium on Multimedia, {ISM}
  2010, Taichung, Taiwan, December 13-15, 2010}, pages 296--301, 2010.
\newblock \doi{10.1109/ISM.2010.51}.
\newblock URL \url{http://dx.doi.org/10.1109/ISM.2010.51}.

\bibitem[Joutou and Yanai(2009)]{Joutou:2009:FIR:1818719.1818816}
Taichi Joutou and Keiji Yanai.
\newblock A food image recognition system with multiple kernel learning.
\newblock In \emph{Proceedings of the 16th IEEE International Conference on
  Image Processing}, ICIP'09, pages 285--288, Piscataway, NJ, USA, 2009. IEEE
  Press.
\newblock ISBN 978-1-4244-5653-6.
\newblock URL \url{http://dl.acm.org/citation.cfm?id=1818719.1818816}.

\bibitem[Wu et~al.(2015)Wu, Yan, Shan, Dang, and
  Sun]{DBLP:journals/corr/WuYSDS15}
Ren Wu, Shengen Yan, Yi~Shan, Qingqing Dang, and Gang Sun.
\newblock Deep image: Scaling up image recognition.
\newblock \emph{CoRR}, arXiv:1501.02876, 2015.
\newblock URL \url{http://arxiv.org/abs/1501.02876}.

\bibitem[Wang et~al.(2015)Wang, Liu, Yang, Han, and Huang]{wang2015deep}
Zhaowen Wang, Ding Liu, Jianchao Yang, Wei Han, and Thomas Huang.
\newblock Deep networks for image super-resolution with sparse prior.
\newblock In \emph{Proceedings of the IEEE International Conference on Computer
  Vision}, pages 370--378, 2015.

\bibitem[{Jin} et~al.(2016){Jin}, {Chen}, {Dong}, {Feng}, and
  {Yan}]{2016arXiv160705440J}
X.~{Jin}, Y.~{Chen}, J.~{Dong}, J.~{Feng}, and S.~{Yan}.
\newblock {Collaborative Layer-wise Discriminative Learning in Deep Neural
  Networks}.
\newblock \emph{ArXiv e-prints}, July 2016.

\bibitem[F{\"{u}}rnkranz and Joachims(2010)]{DBLP:conf/icml/2010}
Johannes F{\"{u}}rnkranz and Thorsten Joachims, editors.
\newblock \emph{Proceedings of the 27th International Conference on Machine
  Learning (ICML-10), June 21-24, 2010, Haifa, Israel}, 2010. Omnipress.

\bibitem[Deng et~al.(2009)Deng, Dong, Socher, Li, Li, and
  Fei-Fei]{imagenet_cvpr09}
J.~Deng, W.~Dong, R.~Socher, L.-J. Li, K.~Li, and L.~Fei-Fei.
\newblock {ImageNet: A Large-Scale Hierarchical Image Database}.
\newblock In \emph{CVPR09}, 2009.

\end{thebibliography}

\end{document}